\documentclass[letterpaper, 10 pt, conference]{class/ieeeconf}  

\IEEEoverridecommandlockouts                              
\usepackage{xcolor}
\usepackage{siunitx}  
\usepackage{graphicx}
 \graphicspath{./figures/robot_demonstration}

\usepackage{amsmath}
\usepackage{amssymb}
\usepackage{latexsym}
\usepackage{url}
\usepackage{cite}
\usepackage{relsize}
\usepackage{multirow}
\usepackage{afterpage}
\usepackage{ifthen}
\usepackage{graphicx}
\usepackage{algpseudocode,algorithm,algorithmicx}
\usepackage{subfigure}
\usepackage{flushend}
\usepackage{epstopdf}
\usepackage{stackengine}
\usepackage{textcomp} 

\usepackage{gensymb}
\usepackage{booktabs}
\usepackage{makecell}
\usepackage{hhline}
\usepackage{courier}
\usepackage{lipsum}

\usepackage{xcolor} 
            
\makeatletter
\let\NAT@parse\undefined
\makeatother
\usepackage[bookmarks=false,linkcolor=blue, urlcolor=blue, citecolor=blue]{hyperref} 

\hypersetup{
    colorlinks=true,
    linkcolor=red,
    filecolor=magenta,      
    urlcolor=blue,
    pdfstartview={FitH},
    citecolor =blue
    }

\usepackage{xspace}

\title{\LARGE \bf Robotic-CLIP: Fine-tuning CLIP on Action Data for \\ Robotic Applications }

\author{Nghia Nguyen$^{1}$, Minh Nhat Vu$^{2,3}$, Tung D. Ta$^4$, Baoru Huang$^7$, Thieu Vo$^5$, Ngan Le$^6$,  Anh Nguyen$^7$
\thanks{$^1$ FPT Software AI Center, Vietnam {\tt nghiant100@fpt.com}}
\thanks{$^2$ Automation \& Control Institute (ACIN), TU Wien, Vienna, Austria}
\thanks{$^3$ Center for Vision, Automation \& Control, AIT Austrian Institute of Technology (GmbH), Vienna, Austria}
\thanks{$^4$ The University of Tokyo, Japan}
\thanks{$^5$ National University of Singapore, Singapore}
\thanks{$^6$ Department of Computer Science \& Computer Engineering, University of Arkansas, USA}
\thanks{$^7$ Department of Computer Science, University of Liverpool, UK}}

\begin{document}

\newtheorem{problem}{Problem}
\newtheorem{lemma}{Lemma}
\newtheorem{theorem}[lemma]{Theorem}
\newtheorem{claim}{Claim}
\newtheorem{corollary}[lemma]{Corollary}
\newtheorem{definition}[lemma]{Definition}
\newtheorem{proposition}[lemma]{Proposition}
\newtheorem{remark}[lemma]{Remark}
\newenvironment{LabeledProof}[1]{\noindent{\it Proof of #1: }}{\qed}

\def\beq#1\eeq{\begin{equation}#1\end{equation}}
\def\bea#1\eea{\begin{align}#1\end{align}}
\def\beg#1\eeg{\begin{gather}#1\end{gather}}
\def\beqs#1\eeqs{\begin{equation*}#1\end{equation*}}
\def\beas#1\eeas{\begin{align*}#1\end{align*}}
\def\begs#1\eegs{\begin{gather*}#1\end{gather*}}

\newcommand{\poly}{\mathrm{poly}}
\newcommand{\eps}{\epsilon}
\newcommand{\e}{\epsilon}
\newcommand{\polylog}{\mathrm{polylog}}
\newcommand{\rob}[1]{\left( #1 \right)} 
\newcommand{\sqb}[1]{\left[ #1 \right]} 
\newcommand{\cub}[1]{\left\{ #1 \right\} } 
\newcommand{\rb}[1]{\left( #1 \right)} 
\newcommand{\abs}[1]{\left| #1 \right|} 
\newcommand{\zo}{\{0, 1\}}
\newcommand{\zonzo}{\zo^n \to \zo}
\newcommand{\zokzo}{\zo^k \to \zo}
\newcommand{\zot}{\{0,1,2\}}
\newcommand{\en}[1]{\marginpar{\textbf{#1}}}
\newcommand{\efn}[1]{\footnote{\textbf{#1}}}
\newcommand{\vecbm}[1]{\boldmath{#1}} 
\newcommand{\uvec}[1]{\hat{\vec{#1}}}
\newcommand{\thv}{\vecbm{\theta}}
\newcommand{\junk}[1]{}
\newcommand{\var}{\mathop{\mathrm{var}}}
\newcommand{\rank}{\mathop{\mathrm{rank}}}
\newcommand{\diag}{\mathop{\mathrm{diag}}}
\newcommand{\tr}{\mathop{\mathrm{tr}}}
\newcommand{\acos}{\mathop{\mathrm{acos}}}
\newcommand{\atantwo}{\mathop{\mathrm{atan2}}}
\newcommand{\SVD}{\mathop{\mathrm{SVD}}}
\newcommand{\quadf}{\mathop{\mathrm{q}}}
\newcommand{\linterp}{\mathop{\mathrm{l}}}
\newcommand{\sgn}{\mathop{\mathrm{sign}}}
\newcommand{\sym}{\mathop{\mathrm{sym}}}
\newcommand{\avg}{\mathop{\mathrm{avg}}}
\newcommand{\mean}{\mathop{\mathrm{mean}}}
\newcommand{\erf}{\mathop{\mathrm{erf}}}
\newcommand{\grad}{\nabla}
\newcommand{\R}{\mathbb{R}}
\newcommand{\defeq}{\triangleq}
\newcommand{\dims}[2]{[#1\!\times\!#2]}
\newcommand{\sdims}[2]{\mathsmaller{#1\!\times\!#2}}
\newcommand{\udims}[3]{#1}
\newcommand{\udimst}[4]{#1}
\newcommand{\com}[1]{\rhd\text{\emph{#1}}}
\newcommand{\ind}{\hspace{1em}}
\newcommand{\argmin}[1]{\underset{#1}{\operatorname{argmin}}}
\newcommand{\floor}[1]{\left\lfloor{#1}\right\rfloor}
\newcommand{\step}[1]{\vspace{0.5em}\noindent{#1}}
\newcommand{\quat}[1]{\ensuremath{\mathring{\mathbf{#1}}}}
\newcommand{\norm}[1]{\left\lVert#1\right\rVert}
\newcommand{\ignore}[1]{}
\newcommand{\specialcell}[2][c]{\begin{tabular}[#1]{@{}c@{}}#2\end{tabular}}
\newcommand*\Let[2]{\State #1 $\gets$ #2}
\newcommand{\algorithmicbreak}{\textbf{break}}
\newcommand{\Break}{\State \algorithmicbreak}
\newcommand{\ra}[1]{\renewcommand{\arraystretch}{#1}}

\renewcommand{\vec}[1]{\mathbf{#1}} 

\algdef{S}[FOR]{ForEach}[1]{\algorithmicforeach\ #1\ \algorithmicdo}
\algnewcommand\algorithmicforeach{\textbf{for each}}
\algrenewcommand\algorithmicrequire{\textbf{Require:}}
\algrenewcommand\algorithmicensure{\textbf{Ensure:}}
\algnewcommand\algorithmicinput{\textbf{Input:}}
\algnewcommand\INPUT{\item[\algorithmicinput]}
\algnewcommand\algorithmicoutput{\textbf{Output:}}
\algnewcommand\OUTPUT{\item[\algorithmicoutput]}

\maketitle
\thispagestyle{empty}
\pagestyle{empty}

\begin{abstract}
Vision language models have played a key role in extracting meaningful features for various robotic applications. Among these, Contrastive Language-Image Pretraining (CLIP) is widely used in robotic tasks that require both vision and natural language understanding. However, CLIP was trained solely on static images paired with text prompts and has not yet been fully adapted for robotic tasks involving dynamic actions. In this paper, we introduce Robotic-CLIP to enhance robotic perception capabilities. We first gather and label large-scale action data, and then build our Robotic-CLIP by fine-tuning CLIP on $309,433$ videos ($\approx7.4$ million frames) of action data using contrastive learning. By leveraging action data, Robotic-CLIP inherits CLIP's strong image performance while gaining the ability to understand actions in robotic contexts. Intensive experiments show that our Robotic-CLIP outperforms other CLIP-based models across various language-driven robotic tasks. Additionally, we demonstrate the practical effectiveness of Robotic-CLIP in real-world grasping applications.


\end{abstract}


\section{INTRODUCTION} \label{Sec:Intro}
Recent advancements in Vision Language Models (VLMs) have opened up exciting possibilities for improving robotic perception and control~\cite{brohan2023rt,shah2023lm,radosavovic2023real, ma2023liv}. Leveraging the capabilities of VLMs, researchers have explored novel approaches to enhance robotic understanding of the environment~\cite{chen2023clip2scene,vuong2024language,shridhar2022cliport}. Notably, the RT-2 model~\cite{brohan2023rt} employs Large Language Models (LLMs) as vision language action models to control the robots. Octo~\cite{team2024octo} was proposed as a generalist robot policy. In~\cite{shah2023lm}, a VLM was used to infer a joint probability distribution over textual landmarks and image observations for the robot navigation task. Not only being used for 2D images, VLMs are also used to enhance the spatial reasoning capabilities of the robots in 3D space~\cite{chen2024spatialVLM}. Other models~\cite{huang2022inner,sermanet2024robovqa} enable grounded closed-loop feedback for robot planning with VLMs. All of these studies contribute to improving the human-robot interaction.

\begin{figure}
\centering
\includegraphics[width=0.99\linewidth, height=0.75\linewidth]{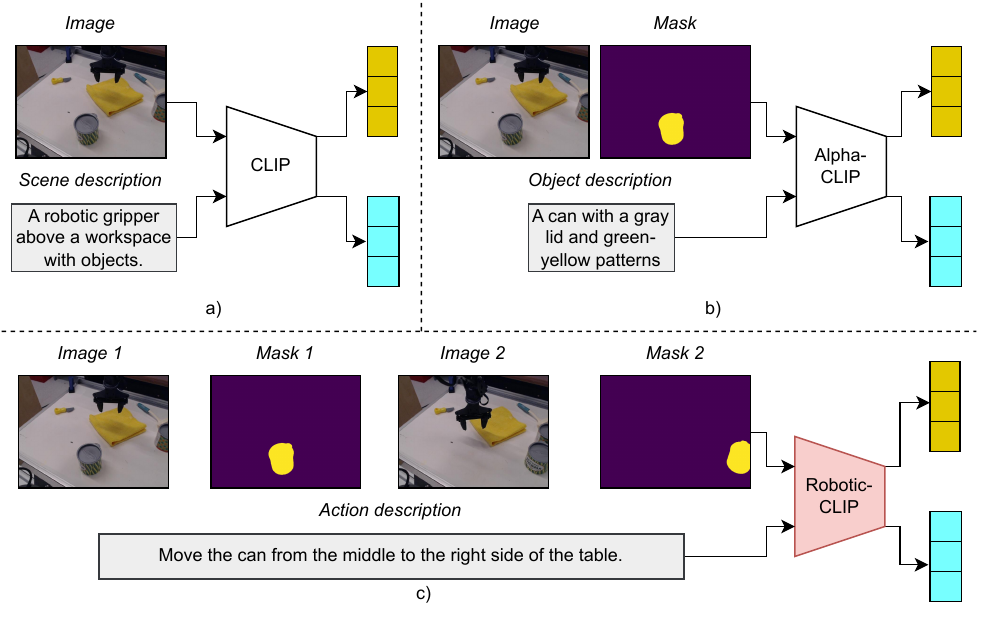}
\vspace{0ex}
\caption{The comparison between: (a) CLIP~\cite{radford2021learning}, (b) Alpha-CLIP~\cite{sun2024alpha}, and (c) our Robotic-CLIP. CLIP aligns scene descriptions with images, while Alpha-CLIP adds object masks. Our method aligns paired frames with action descriptions and masks to capture the action in robotic tasks.}
\label{fig:intro}
\end{figure}

Currently, VLMs are trained on large-scale datasets containing a substantial number of images and text prompts~\cite{radford2021learning,li2023blip}. They have demonstrated effectiveness during validation across various tasks, including both seen and unseen scenarios. However, while VLMs excel at processing static images, they struggle with time-dependent data such as videos or motion sequences~\cite{radosavovic2023real}. Recent works have aligned text and video, but they often rely on independent frames without capturing the necessary temporal dependencies, making it challenging to embed critical robot actions~\cite{karamcheti2023language}. The lack of effective action modelling can hinder the ability to predict the robots' future perception or control states. Moreover, since robots can perform diverse actions, fine-tuning VLMs to understand the robotic action is a crucial task.

Among VLMs, the Contrastive Language Image Pre-training (CLIP) is widely used in robotic tasks that require vision and language understanding~\cite{tao2023galip}. CLIP has demonstrated superior performance compared to previous uni-modal models~\cite{dosovitskiy2020image,devlin2018bert}. Numerous works have utilized CLIP as a feature extractor for robot tasks such as language-driven grasp detection~\cite{vuong2023grasp}, zero-shot detection~\cite{liu2023grounding}, and language navigation~\cite{shah2023lm, huang2023visual}. Recently, several works have explored fine-tuning CLIP with different types of data for specific tasks. Beyond RGB images and text in the original CLIP model, Alpha-CLIP~\cite{sun2024alpha} was proposed to include a mask channel, Prompt-CLIP used prompting techniques for depth images~\cite{auty2023learning}, or Point-CLIP was fine-tuned on 3D point clouds for 3D understanding tasks~\cite{zhang2022pointclip}. In~\cite{sun2024alpha,zhang2022pointclip}, the fine-tuned CLIP models have shown promising results on specific data. However, none of them have explored the relationship between images and the actions described in the text prompt for robotic applications. This is a common issue with CLIP and its variants, as they are trained on \textit{static data} rather than \textit{action data}, hence they may not fully perceive action-related information for robotic applications.


In this paper, we introduce Robotic-CLIP, a new feature extractor that not only semantically aligns video frames with text instructions but also effectively captures and highlights actions within the video. The overview of our approach is illustrated in Fig.~\ref{fig:intro}. Unlike~\cite{radford2021learning,sun2024alpha}, which focus mainly on static scene or object recognition, our method introduces the use of two distinct frames to better capture and interpret actions. This enables the model to understand action relationships better, enhancing its overall performance in robotic tasks. Our primary objective is to fine-tune the CLIP model for robotic applications, ensuring robust performance across various downstream tasks. Building upon CLIP, our approach enhances the generalizability of representation vectors derived from large-scale text-video pairs. To further augment the model's knowledge, we propose a dataset generalization pipeline that leverages foundation models to autonomously label large-scale action data for fine-tuning our model~\cite{liu2023grounding, kirillov2023segment, akbik2019flair}. Additionally, we introduce a fine-tuning method that enables the model to learn new action concepts while avoiding catastrophic forgetting of previously acquired knowledge from the original CLIP. Our approach enables zero-shot learning, allowing robots to tackle diverse tasks without compromising performance. We verify our Robotic-CLIP on several language-driven robotic tasks and also in real-world robotic experiments.

Our contributions are summarized as follows:
\begin{itemize}
    \item 
    We introduce Robotic-CLIP, a simple yet effective model designed for language-driven robotic tasks.
    \item
    We propose a pipeline for generating large-scale action data and a new fine-tuning technique to enable the model to learn deeper action understanding.
    \item 
    We conduct intensive experiments across different robotic tasks to validate our model performance. Our code and models will be released.
    
\end{itemize}
\section{RELATED WORK} \label{Sec:rw}


\textbf{Vision Language Representations for Robotics.} Recently, many works have studied human-robot interaction from a language-driven perspective to enhance performance. R3M~\cite{nair2022r3m}, LIV~\cite{ma2023liv} have demonstrated their effectiveness in extracting important information for robotic tasks. Furthermore, these language-driven approaches have paved the way for more seamless communication between humans and robots. By leveraging natural language processing techniques, robots can understand and respond to user commands, queries, and instructions in a more intuitive manner. In~\cite{nair2022r3m}, Nair \textit{et al.} leverage attention mechanism allows it to focus on relevant parts of the input text, enabling precise information extraction. Similarly, Ma \textit{et al.} use a framework that learns embedding vectors based on Value Implicit Pre-training (VIP)~\cite{ma2022vip} and CLIP~\cite{radford2021learning}. 
Some other studies learn to perform masked reconstruction from images to understand the contexts, such as MVP~\cite{radosavovic2023real} and Voltron~\cite{karamcheti2023language}. Despite their advancements, the robot datasets used for these methods are often collected from a small number of task demonstrations, rather than from real objects in real-world contexts.

\textbf{CLIP in Robotic Applications.} Language-driven robotic tasks require both visual and textual information, making CLIP~\cite{tao2023galip} an ideal choice. The pick-and-place problem is common in robotics, with the two related tasks being grasp detection~\cite{kumra2020antipodal, ainetter2021end, xu2023joint, morrison2018closing, vuong2024language, shridhar2022cliport} and affordance detection~\cite{nguyen2023open,van2023open, cheraghian2020transductive}. Shridhar \textit{et al.} proposed a two-stream architecture with semantic and spatial pathways for vision-based manipulation~\cite{shridhar2022cliport}. To improve performance on unseen datasets, the author in~\cite{xu2023joint} proposed a new method that is more flexible in handling language instructions and is not limited by visual grounding errors. For real-world grasp detection, Vuong \textit{et al.} introduced the Grasp-Anything dataset~\cite{vuong2023grasp} and proposed a method to generate grasp poses by leveraging the power of the diffusion model~\cite{vuong2024language}. To learn different types of data in multi-modal scenarios, a variety of CLIP variants have emerged. Some of these variants allow region awareness in specific defined areas, enabling a focus on important information within images. MaskCLIP~\cite{dong2023maskclip} and ODISE~\cite{xu2023open} use attention masks to make CLIP focus more on local regions. AlphaCLIP~\cite{sun2024alpha} creates an additional alpha channel for learning semantic detail at wherever in image. Other models leverage the power of CLIP and are fine-tuned on different types of data. Guzhou \textit{et al.} proposed AudioCLIP\cite{guzhov2022audioclip}, enabling a tri-modal hybrid architecture for aligning image, text, and audio data. PointCLIP~\cite{zhang2022pointclip} projects 3D point clouds to multiple 2D depth maps and uses an inner-view adapter to finetune models. Although they obtain good performance, these models do not learn action information, which is essential for tasks involving dynamic interactions. Their primary focus is on static object recognition or localized regions, limiting their ability to capture the temporal and sequential aspects of actions. 

\textbf{Contrastive Learning for Robotics.} Robots increasingly require learning knowledge as similarly to humans as possible, and we usually need to train robot models from different types of data. Contrastive learning is a popular and effective solution for multi-modal learning problems. PointCLIP~\cite{zhang2022pointclip} is a variant of the CLIP~\cite{radford2021learning} model that enables zero-shot classification of 3D objects. Wang~\textit{et al.} propose an effective few-shot point cloud semantic segmentation approach~\cite{wang2023few}. To extend the affordance detection task, the authors in ~\cite{nguyen2023open, van2023open} align open vocabulary affordance labels with individual points in the 3D point cloud in a zero-shot manner using contrastive loss. Vuong~\textit{et al.}~\cite{vuong2024language} also use contrastive loss to guide the grasp pose output, ensuring it aligns correctly with the text and image input. For reinforcement learning tasks, CoDER~\cite{zhan2022learning} leverages contrastive pre-training and data augmentation to initialize encoders and improve efficiency. Xing~\textit{et al.}~\cite{xing2024contrastive} propose an adaptive contrastive learning method to enhance scene transfer, enabling zero-shot deployment across unseen environments. In this study, unlike previous methods that only focus on aligning individual images with text~\cite{radford2021learning,sun2024alpha,vuong2023grasp}, we align the text with the video frames describing actions, ensuring the similarity with the text increases over time to better capture the action. This more effectively handles temporal changes, which previous methods have not addressed.

\begin{figure*}[h]
	\centering
\includegraphics[width=1.0\linewidth, height = 0.45\linewidth]{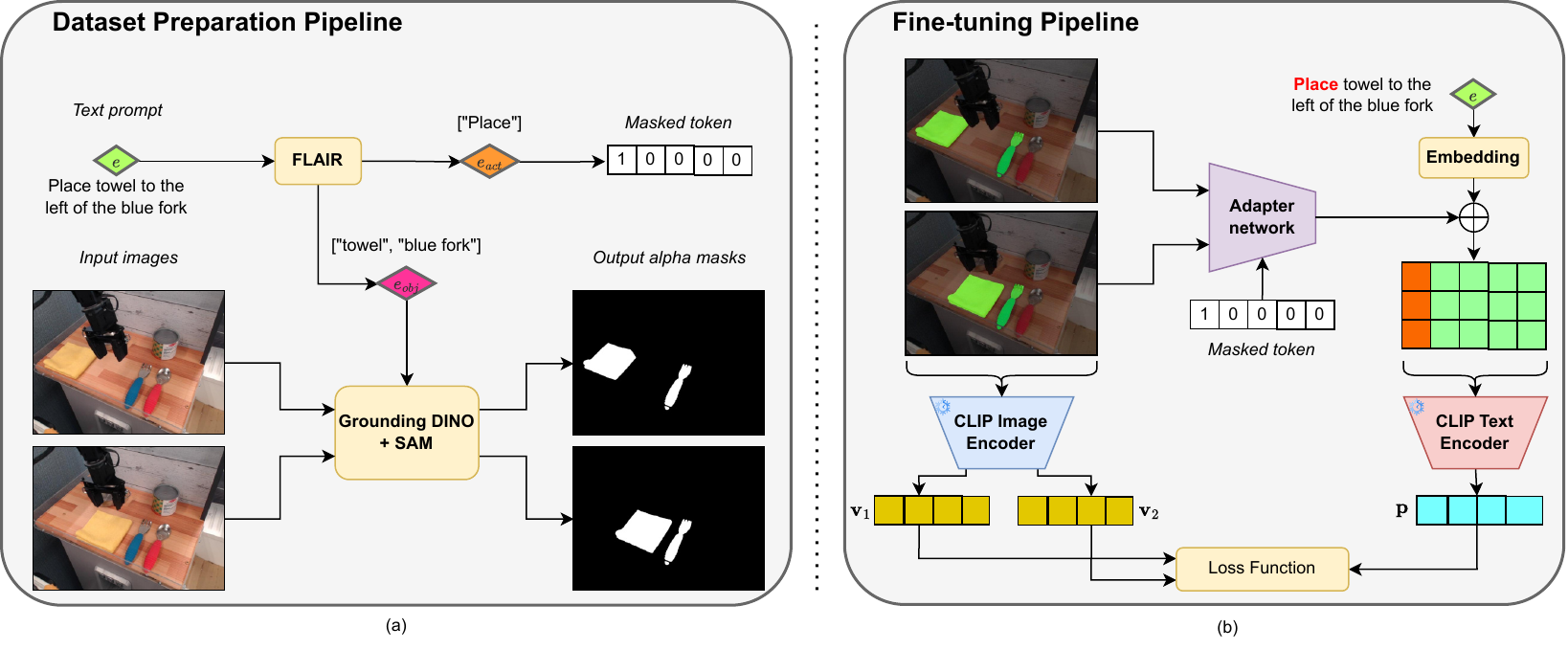}
 \vspace{-1ex}
 \caption{The dataset preparation pipeline (a) and our fine-tuning pipeline (b).}
 \vspace{1ex}
\label{fig:finetune_technique}
\end{figure*} 
\section{METHODOLOGY} \label{Sec:method}
\subsection{Motivation}
The original CLIP model is commonly used in various tasks and is trained from a large dataset of approximately 400M image-text pairs. This extensive sample size allows CLIP to achieve impressive performance on static images across multiple tasks~\cite{liu2023clip}. However, CLIP is only trained on static text-image pairs and lacks of knowledge about action understanding for robotic applications. To address this issue, we collect large-scale action data and propose a triplet loss to differentiate the semantic behaviors of robots, enabling better alignment between robot actions and text descriptions.

\subsection{Data Preparation}
To fine-tune the CLIP model with videos and prompts describing robot actions, we first extract all frames from the videos of three large-scale action datasets, including Something-Something V2~\cite{goyal2017something}, BridgeData V2~\cite{walke2023bridgedata}, and RoboSet~\cite{bharadhwaj2023roboagent} (Table~\ref{tab_dataset}). All video frames are preprocessed to create the alpha masks containing the corresponding objects in the text. Our data preparation pipeline consists of the following two steps as shown in Fig.~\ref{fig:finetune_technique} (a).

\textbf{Query extraction.} Given an input prompt instruction, we first extract the object and action list from the text prompts described the videos. The raw videos and prompt instructions are utilized from datasets of robot manipulation behaviors and human object interaction~\cite{walke2023bridgedata, bharadhwaj2023roboagent, goyal2017something}. Table~\ref{tab_dataset} shows the statistics of three large-scale datasets that we use to prepare the data. We employ FLAIR~\cite{akbik2019flair} to automatically extract part-of-speech (POS) tags for nouns and verbs in the prompt. Nouns represent objects, while verbs represent actions. For example, given the text prompt \textit{``Place towel to the left of the blue fork”}, we extract the relevant objects as [\textit{``towel”}, \textit{``blue fork”}] for the next step, and the action \textit{``place”} will be used during the model’s fine-tuning.

\begin{table}[h]
\caption{The datasets used for our method}
\label{tab_dataset}
\vskip 0.15in
\setlength{\tabcolsep}{5pt}
\renewcommand{\arraystretch}{1.2}

\begin{center}
\begin{tabular}{lrcc}
\toprule
Dataset & \#Videos & \#Action Categories\\
\midrule
Something-Something V2~\cite{goyal2017something} & 220,847 & 174\\
BridgeData V2~\cite{walke2023bridgedata} & 60,096 & 13\\
RoboSet~\cite{bharadhwaj2023roboagent} & 28,500 & 12\\
\bottomrule
\textbf{Total} & 309,443 & 199\\
\end{tabular}
\end{center}
\vskip -0.1in
\end{table}

\textbf{Object masking.} For each action video, we extract all frames from the video~\cite{wang2024videocomposer}. The objects in the text prompts are then used to create masks for all frames. We employ GroundingDINO~\cite{liu2023grounding} and SAM~\cite{kirillov2023segment} to segment the objects from the images. This process is performed for all the frames. The resulting masks help the model focus on the important regions within the images. Overall, we have processed $309,433$ videos ($\approx 7.4$ million frames) of robot and human actions and corresponding text prompts. The sufficiently large and diverse data allows us to fine-tune the CLIP model effectively while ensuring data quality and the corresponding mask-guided actions.

\subsection{Robotic-CLIP}
Given a dataset with action videos paired with text descriptions for robotic applications, each video contains a sequence of RGBA frames $[\mathbf{I}_1, \mathbf{I}_2, \dots, \mathbf{I}_N]$, $\mathbf{I}_t\in\mathbb{R}^{H\times W\times 4}$. Each language prompt $\mathbf{e} \in \mathbb{R}^{L\times D}$ represents a sequence of $L$ tokens, where each token is embedded into a $D$-dimensional vector space. Our goal is to learn an image encoder $\mathbf{f}_{\text{image}}(\mathbf{I}_t)$ and a text encoder $\mathbf{f}_{\text{text}}(\mathbf{e})$ that maps images and prompt to a continuous embedding. We randomly choose two frames $\mathbf{I}_{t_1}$ and $\mathbf{I}_{t_2}$ from the video $(t_1<t_2)$ to represent the action. We define output representations:\
$
    \mathbf{v}_1 = \mathbf{f}_{\text{image}}(\mathbf{I}_{t_1})
$, $
    \mathbf{v}_2 = \mathbf{f}_{\text{image}}(\mathbf{I}_{t_2})
$, $
    \mathbf{p} = \mathbf{f}_{\text{text}}(\mathbf{e})
$. Once trained, we can reuse $\mathbf{f}$ for downstream tasks. Specifically, we utilize our pretrained model as a feature representation for a specific robotic task.

\textbf{Model structure.} Our Robotic-CLIP builds on the CLIP~\cite{tao2023galip} and Alpha-CLIP~\cite{sun2024alpha} architecture, consisting of a text encoder and an image encoder. The image encoder processes two inputs: an RGB image representing the scene and a binary mask highlighting relevant objects. The output includes vector representations for both RGBA images and the text prompt. As shown in Fig.~\ref{fig:finetune_technique} (b), we fine-tune the model using two RGBA images, one before and one after the action. To prevent catastrophic forgetting, both the image encoder and text encoder from the pretrained Alpha-CLIP are frozen. We introduce an Adapter Network to map the input images into the text space, ensuring effective alignment of visual and textual representations while retaining the model’s prior knowledge.

Suppose that we have an learnable Adapter Network $\mathbf{s}_{\phi}(.)$ and binary action masked token $\mathbf{m}=[m_1,m_2,\dots,m_L]$. The text embedding after tuning is modified as follows:
\begin{equation} \mathbf{e}_{\text{action}} = \frac{\mathbf{s}_{\phi}(\mathbf{I}_{t_1}) + \mathbf{s}_{\phi}(\mathbf{I}_{t_2})}{2}
\end{equation}
\begin{equation}
    \mathbf{e}' = \mathbf{e} \odot (1 - \mathbf{m}) + \mathbf{e}_{\text{action}} \odot \mathbf{m}
\end{equation}
\begin{equation}
    \mathbf{p} = \mathbf{f}_{\text{text}}(\mathbf{e}')
\end{equation}
where $\mathbf{e}_{\text{action}} \in \mathbb{R}^{D}$ is the action representation learned by the Adapter Network based on the two input images $\mathbf{I}_{t_1}$ and $\mathbf{I}_{t_2}$. The element-wise product $\odot$ selectively modifies the action-related components of the text embedding while preserving the rest. 


\textbf{Adapter Network.} In practice, we utilize an Adapter Network consisting of a 12-layer Vision Transformer~\cite{dosovitskiy2020image} to extract action embeddings. To accommodate RGBA images, we modify the Linear Projection input layer~\cite{dosovitskiy2020image} to accept four-channel inputs instead of the standard three-channel RGB, enabling the model to process mask information alongside colour data. Unlike previous approaches, which typically place the Adapter Network on top of the CLIP representations~\cite{gao2024clip}, our design places it before the encoders, as shown in Fig.~\ref{fig:finetune_technique} (b). This design choice is motivated by two key reasons: \textit{i)} placing the adapter after the encoders may alter the original embeddings and reduce their effectiveness, and \textit{ii)} our fine-tuning strategy allows focused adjustments to the action embeddings while preserving the integrity of the other components. This ensures the model retains its object recognition capabilities while learning new action-specific details from our datasets.

\textbf{Contrastive Learning for Action Embedding.} 
Given a batch of $B$ videos, we compute the correlation between the embeddings of two frames $\mathbf{I}_{t_1},\mathbf{I}_{t_2}$ from video $i$ and the text prompt $\mathbf{e}$ from video $j$:
\begin{equation}
    F_{i,j}^1 = \frac{\mathbf{v}_{1i}^T\mathbf{p}_j}{\norm{\mathbf{v}_{1i}}\norm{\mathbf{p}_j}}
\end{equation}
\begin{equation}
    F_{i,j}^2 = \frac{\mathbf{v}_{2i}^T\mathbf{p}_j}{\norm{\mathbf{v}_{2i}}\norm{\mathbf{p}_j}}
\end{equation}

The alignment score between each image and its corresponding text is calculated using the softmax function:
\begin{equation}
    S_{i}^1 = \frac{\exp(F_{i,i}^1/\tau)}{\sum_{j=1}^{B}{\exp(F_{i,j}^1/\tau)}}
\end{equation}
\begin{equation}
    S_{i}^2 = \frac{\exp(F_{i,i}^2/\tau)}{\sum_{j=1}^{B}{\exp(F_{i,j}^2/\tau)}}
\end{equation}

The contrastive loss for image-text correlation is:
\begin{equation}
    \mathcal{L}_{\text{contrastive}} = -\frac{1}{B}\sum_{i=1}^{B} {\log{S_{i}^1 
 + \log{S_{i}^2}}}
\end{equation}

To highlight embedding vector for action word in an unique video, we use triple loss for output embedding:
\begin{equation}
\mathcal{L}_{\text{triplet}} = \frac{1}{B}\sum_{i=1}^{B}\max(\norm{\mathbf{v}_{2i} - \mathbf{p}_i} - \norm{\mathbf{v}_{1i}-\mathbf{p}_i} + \epsilon , 0)
\end{equation}
where $\epsilon$ is margin between the positive and negative pairs. The overall training objective for our method is:
\begin{equation}
    \mathcal{L}_{\text{total}} = \lambda\mathcal{L}_{\text{contrastive}} + \mathcal{L}_{\text{triplet}}
\end{equation}
where $\lambda$ is a hyper-parameter to balance the loss. $\lambda$ is set to $0.1$ in our implementation.
\section{Experiments} \label{Sec:exp}

In this section, we perform several downstream robotic tasks to demonstrate the effectiveness of our Robotic-CLIP. Specifically, we use CLIP~\cite{radford2021learning}, Alpha-CLIP~\cite{sun2024alpha}, and Robotic-CLIP as the information encoders for downstream tasks. To ensure fairness and applicability in unseen scenarios, we freeze the pretrained CLIP-based models during training and only retrain the downstream tasks using different CLIP-based models. Additionally, we analyse the triplet loss and provide real-world robotic experiment results.

\subsection{Robotic-CLIP in Language-driven Grasp Detection}
\begin{table}[h]
\caption{Comparison of Language-driven Grasp Detection methods with different CLIP models}
\label{tab_grasp}
\vskip 0.15in
\setlength{\tabcolsep}{4pt}
\begin{center}
\renewcommand{\arraystretch}{1.1}
\begin{tabular}{lccc}
\toprule
\textbf{Method} & \textbf{CLIP}~\cite{radford2021learning} & \textbf{Alpha-CLIP}~\cite{sun2024alpha} & \textbf{Robotic-CLIP} \\
\midrule
GR-ConvNet~\cite{kumra2020antipodal} & 0.24 ($\uparrow$0.12) & 0.31 ($\uparrow$0.05) & 0.36\\
Det-Seg~\cite{ainetter2021end} & 0.20 ($\uparrow$0.08) & 0.23 ($\uparrow$0.05) & 0.28\\
GG-CNN~\cite{morrison2018closing} & 0.10 ($\uparrow$0.07) & 0.14 ($\uparrow$0.03) & 0.17\\
LGD~\cite{vuong2024language}& 0.45 ($\uparrow$0.05) & 0.49 ($\uparrow$0.01) & 0.50\\
LLGD~\cite{nguyen2024lightweight}& 0.46 ($\uparrow$0.05) & 0.50 ($\uparrow$0.01) & 0.51\\
\bottomrule
\end{tabular}
\end{center}
\vskip -0.1in
\end{table}

We select the Grasp-Anything dataset~\cite{vuong2023grasp} for the language-driven grasp detection task. This dataset comprises 1M images, each accompanied by one or several prompts describing grasping actions on specific objects. We select five baseline methods (GR-ConvNet~\cite{kumra2020antipodal}, Det-Seg~\cite{ainetter2021end}, GG-CNN~\cite{morrison2018closing}, LGD~\cite{vuong2024language}, LLGD~\cite{nguyen2024lightweight}) and retrain them using Alpha-CLIP~\cite{sun2024alpha} and our Robotic-CLIP. We note that we only retrain the baselines with different CLIP-based models, and keep the original architectures unchanged. The primary evaluation metric is the IoU~\cite{vuong2023grasp}. 

Table~\ref{tab_grasp} shows the results of different baselines when we integrate our Robotic-CLIP into them. Table~\ref{tab_grasp} shows that our Robotic-CLIP outperforms other CLIP models by a clear margin. This demonstrates that learning the actions represented in the data is useful for the language grasping tasks, hence consequently improving the results of all baselines.

\begin{figure}
\centering
\includegraphics[width=1.0\linewidth, height = 0.85\linewidth]{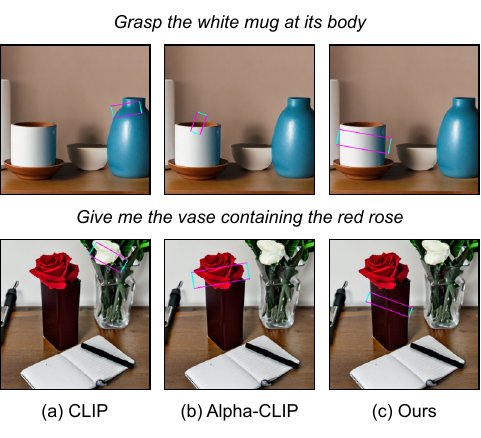}
\caption{Language-drive grasp detection results of LLGD~\cite{nguyen2024lightweight} using different CLIP models.}
\label{fig:visualize_grasp}
\end{figure}

Fig.~\ref{fig:visualize_grasp} presents a comparison of different CLIP models for the language-driven grasp detection task. The results indicate that Robotic-CLIP generates more semantically appropriate grasp poses based on the given text queries compared to CLIP~\cite{radford2021learning} and Alpha-CLIP~\cite{sun2024alpha}. Additionally, our Robotic-CLIP demonstrates a better understanding of complex grasp instructions, such as those involving relative clauses.

\subsection{Robotic-CLIP in Policy Learning}

\begin{table}[h]
\caption{Comparison of Language-Conditioned Policy Learning Methods with Different Models}
\label{tab_manipulation}
\vskip 0.15in
\setlength{\tabcolsep}{5pt}
\begin{center}
\begin{tabular}{lc}
\toprule
Model & Success Rate$\uparrow$  \\
\midrule
LOREL~\cite{nair2022learning} & 9.6$\pm$3.0\\
R3M~\cite{nair2022r3m} & 8.8$\pm$2.7\\
LIV~\cite{ma2023liv} + CLIP~\cite{radford2021learning} & 1.3$\pm$0.8\\
LIV~\cite{ma2023liv} + Alpha-CLIP~\cite{sun2024alpha} & 12.6$\pm$4.5 \\
LIV~\cite{ma2023liv} + Robotic-CLIP (ours) & \textbf{13.3$\pm$4.1} \\
\bottomrule
\end{tabular} 
\end{center}
\vskip -0.1in
\end{table}

In this experiment, we perform policy learning in a multi-task language-conditioned visual manipulation environment. We adopt the setup and dataset from~\cite{nair2022r3m} which used the FrankaKitchen benchmark~\cite{gupta2019relay}. We choose LIV~\cite{ma2023liv} as the baseline model and replace CLIP~\cite{radford2021learning} in its architecture with the pretrained Alpha-CLIP~\cite{sun2024alpha} and our Robotic-CLIP. 

We measure the success rate overall test instances in Table~\ref{tab_manipulation}. The results demonstrate that in the original setup, the LIV~\cite{ma2023liv} + CLIP model fails in almost all cases, while LIV+Alpha-CLIP and LIV+Robotic-CLIP yield significantly better results without requiring a fine-tuning step. This further confirms the effectiveness of our Robotic-CLIP in robotic applications.


\begin{figure}[t]
\centering
\includegraphics[width=0.99\linewidth, height = 0.68\linewidth]{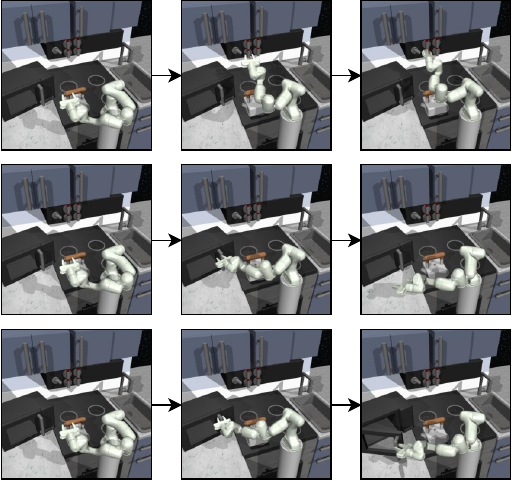}
\vspace{0.5ex}
\caption{Results of policy learning experiments. With the text prompt input ``\textit{Open Microwave}", we compare the robot actions generated by the baseline LIV\cite{ma2023liv} with our method and other CLIP-based models. The top row represents CLIP~\cite{radford2021learning}, the middle row corresponds to Alpha-CLIP~\cite{sun2024alpha}, and the bottom row illustrates our proposed Robotic-CLIP method.}
\label{fig:policy_learning}
\end{figure}

Fig.~\ref{fig:policy_learning} shows the results of policy learning task in the FrankaKitchen environment~\cite{gupta2019relay} using LIV~\cite{ma2023liv} baseline and CLIP-based models. We observe significant differences when using the baseline LIV~\cite{ma2023liv} with different CLIP-based encoders. CLIP\cite{radford2021learning} struggles to accurately identify and interact with the target objects. Alpha-CLIP\cite{sun2024alpha} performs better in recognizing object positions but fails to execute the actions with sufficient precision. In contrast, our Robotic-CLIP completes the task successfully.

\subsection{Robotic-CLIP in Robot Navigation}
\begin{table}[h]
\caption{Comparison of Visual Robot Navigation Methods with Different CLIP-based Models}
\label{tab_navigation}
\vskip 0.15in
\setlength{\tabcolsep}{5pt}
\begin{center}
\begin{tabular}{lcc}
\toprule
Model & Net Success$\uparrow$ \\
\midrule
GPS-Nav~\cite{shah2023lm} + CLIP~\cite{radford2021learning} & 0.22 \\
GPS-Nav~\cite{shah2023lm} + Alpha-CLIP~\cite{sun2024alpha} & 0.28 \\
GPS-Nav~\cite{shah2023lm} + Robotic-CLIP (ours)& \textbf{0.31} \\
\midrule
LM-Nav~\cite{shah2023lm} + CLIP~\cite{radford2021learning} & 0.75 \\
LM-Nav~\cite{shah2023lm} + Alpha-CLIP~\cite{sun2024alpha} & 0.83 \\
LM-Nav~\cite{shah2023lm} + Robotic-CLIP (ours)& \textbf{0.86}\\
\bottomrule
\end{tabular} 
\end{center}
\vskip -0.1in
\end{table}

We conduct the visual language navigation task on the CARLA simulator\cite{dosovitskiy2017carla}. The baseline methods used are LM-Nav~\cite{shah2023lm} and GPS-Nav~\cite{shah2023lm}. Unlike the traditional navigation methods that rely solely on visual information, the visual language navigation task relies on the text instructions (such as \textit{``turn right at the stop sign"}) to guide the robot. In our experiments, CLIP\cite{radford2021learning}, Alpha-CLIP\cite{sun2024alpha}, and Robotic-CLIP are used as the pretrained model for all baselines. We use the planning success metric~\cite{shah2023lm} to compare how effectively each baseline supports robot navigation.

Table~\ref{tab_navigation} shows that our Robotic-CLIP consistently outperforms CLIP and Alpha-CLIP across baseline methods. This demonstrates the robustness and enhanced capability of our proposed Robotic-CLIP in supporting the visual language robot navigation task.

\subsection{Ablation Study}



\textbf{Triplet Loss Analysis.} Table~\ref{tab_triplet_analysis} provides a comparison of the robot policy learning task with and without using triplet loss during the fine-tuning process. The results demonstrate the impact of triplet loss on success rates across various policy learning tasks. This table shows a consistent improvement when triplet loss is used to fine-tune our Robotic-CLIP. This highlights its effectiveness in enhancing action understanding, particularly in complex action scenarios.

\begin{figure}[t]
\centering
\includegraphics[width=1.0\linewidth]{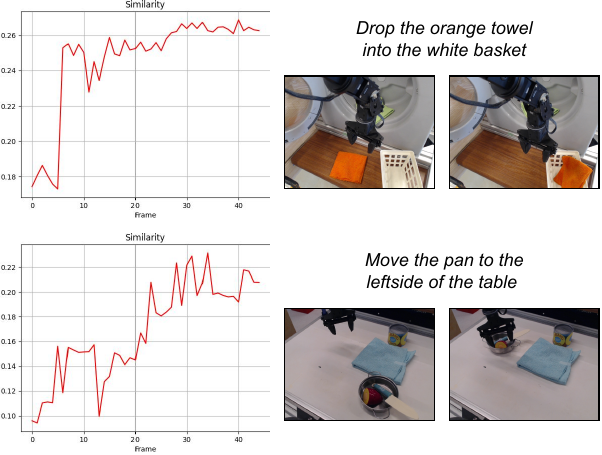}
\vspace{-2ex}
\caption{Text-frame similarity analysis.}
\vspace{2ex}
\label{fig:similarity_analysis}
\end{figure}

\begin{table}[t]
\caption{TRIPLET LOSS ANALYSIS FOR POLICY LEARNING TASK}
\label{tab_triplet_analysis}
\vskip 0.15in
\setlength{\tabcolsep}{5pt}
\begin{center}
\begin{tabular}{lcc}
\toprule
Task & Without triplet loss & With triplet loss \\
\midrule
Open Microwave & 10.4 & 12.7\\
Open left door & 9.8 & 12.3\\
Slide Cabinet & 13.2 & 14.2\\
Switch on Light & 14.1 & 14.5\\
Turn on Stove & 12.5 & 12.8\\
\bottomrule
\end{tabular} 
\end{center}
\vskip -0.1in
\end{table}

\textbf{Text-Frames Alignment Analysis. }To assess our model’s ability to align textual action descriptions with visual frames over time, we perform similarity analysis between the text and the frame sequences. Fig.~\ref{fig:similarity_analysis} presents the similarity trends across various tasks, showing how the alignment between the action descriptions and the visual frames evolves as the action unfolds. In each case, the similarity score increases as the frames progress, demonstrating our model’s effectiveness in capturing the dynamic relationship between the textual instructions and the corresponding visual changes. 


\subsection{Robotic Experiment}

\begin{figure}[h]
\centering
\includegraphics[width=1.0\linewidth]{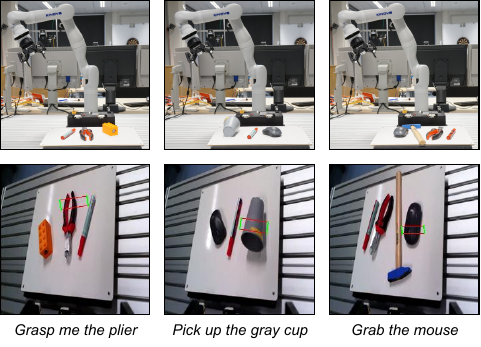}
\vspace{-2ex}
\caption{Robot experiment results with Robotic-CLIP and LLGD~\cite{nguyen2024lightweight}.}
\label{fig:robot}
\vspace{2ex}
\end{figure}

\textbf{Robotic Setup.} We integrate our Robotic-CLIP into a lightweight language-driven grasp detection pipeline LLGD~\cite{nguyen2024lightweight} to form a grasping framework. The Kinova Gen3 7-DoF robot is used to deliver quantifiable outcomes (Fig.~\ref{fig:robot}). Utilization of the RealSense D410 camera enables the translation of grasping information from LLGD into a 6DoF grasp posture, bearing resemblance to~\cite{kumra2020antipodal}. Subsequently, a trajectory optimization planner~\cite{vu2023machine} is used to execute the grasping action. Experiments were conducted on a table surface for the single object scenario and the cluttered scene scenario, wherein various objects were placed randomly. Table~\ref{tab_robotic_results} shows the success rate of our method and other baselines. We can see that our method outperforms other baselines in both single object and cluttered scenarios.

\begin{table}[t]
    \centering
    \caption{\label{table: real-robot-exp} Robotic language-driven grasp detection results}
    \vspace{2ex}
    \renewcommand
\tabcolsep{4pt}
\hspace{1ex}
    \begin{tabular}{lcc}
\toprule
Baseline & Single &  Cluttered\cr 
\midrule
LLGD~\cite{nguyen2024lightweight} + CLIP~\cite{radford2021learning} & 0.42 & 0.40 \\
LLGD~\cite{nguyen2024lightweight} + Alpha-CLIP~\cite{sun2024alpha} & 0.44& 0.41 \\
LLGD~\cite{nguyen2024lightweight} + Robotic-CLIP (ours) & \textbf{0.53} & \textbf{0.50} \\
\bottomrule
\end{tabular}
\label{tab_robotic_results}
\end{table}

\subsection{Limitation}
Despite achieving strong performance in various robotic applications, our Robotic-CLIP still has certain limitations. One of the primary limitations lies in the fact that Robotic-CLIP is fine-tuned solely on 2D video data, which restricts its application in tasks that require 3D spatial understanding, such as those involving point clouds or depth perception. Addressing this limitation would require incorporating 3D data into the training process. This enhancement could enable Robotic-CLIP to better understand the full 3D scenes, opening up possibilities for a broader range of robotic tasks that require precise 3D perception.



\section{Discussion}\label{Sec:con}
We propose Robotic-CLIP, a new model that enhances a robot's ability to comprehend vision language understanding via action data. Unlike other approaches, Robotic-CLIP can capture dynamic information such as actions and movements. The experimental results show that Robotic-CLIP outperforms other CLIP-based methods by a clear margin. We see several interesting future research directions. First, we could extend the problem to handle time-dependent 3D data such as 3D point clouds. This would allow the model to learn cognitive abilities in real-world environments. Second, aligning other types of data beyond images and text, such as robot kinematic or tactile could further enhance the model’s capability to process multi-modal information and open up possibilities for more complex robotic applications. Our code and model will be released.




\bibliographystyle{class/IEEEtran}
\bibliography{class/IEEEabrv,class/reference}
   
\end{document}